\documentclass{article}
\usepackage{arxiv}

\usepackage{cite}
\usepackage{amsmath,amssymb,amsfonts}
\usepackage{algorithmic}
\usepackage{graphicx,color}
\usepackage{textcomp}
\usepackage{xcolor}
\usepackage{hyperref}
\hypersetup{hidelinks=true}
\usepackage{algorithm,algorithmic}
\usepackage{tabularx}
\usepackage{subcaption}
\usepackage{colortbl}

\DeclareMathOperator*{\argmin}{\arg\!\min}

\title{A Gradient Boosting Approach for Training Convolutional and Deep Neural Networks\thanks{This paper is under consideration at IEEE Open Journal of Signal Processing}}
\author{
 Seyedsaman Emami \\
  Escuela Politecnica Superior\\
  Universidad Autonoma de Madrid\\
  Ciudad Universitaria de Cantoblanco, Madrid 28049 \\
  \texttt{emami.seyedsaman@uam.es} \\
   \And
 Gonzalo Mart\'{\i}nez-Mu\~noz \\
  Escuela Politecnica Superior\\
  Universidad Autonoma de Madrid\\
  Ciudad Universitaria de Cantoblanco, Madrid 28049 \\
  \texttt{gonzalo.martinez@uam.es} \\
}

\begin{document}
\maketitle
\begin{abstract}
Deep learning has revolutionized the computer vision and image classification
domains. In this context Convolutional Neural Networks (CNNs) based architectures
are the most widely applied models. In this article, we introduced two
procedures for training Convolutional Neural Networks (CNNs) and Deep Neural
Network based on Gradient Boosting (GB), namely GB-CNN and GB-DNN. These
models are trained to fit the gradient of the loss function or pseudo-residuals
of previous models. At each iteration, the proposed method adds one dense layer
to an exact copy of the previous deep NN model. The weights of the dense layers
trained on previous iterations are frozen to prevent over-fitting, permitting
the model to fit the new dense as well as to fine-tune the convolutional layers
(for GB-CNN) while still utilizing the information already learned. Through
extensive experimentation on different 2D-image classification and tabular
datasets, the presented models show superior performance in terms of classification
accuracy with respect to standard CNN and Deep-NN with the same architectures.
\end{abstract}


The well-known deep learning technique designs a framework using Artificial
Neural Network systems (ANNs). The concept of deep learning after introducing
the AlexNet~\cite{NIPS2012_4824} model gained even more attention.
A high variety of architectures and topologies of deep network models can be
constructed by combining different layer types in the model. Likewise,
Convolutional Neural Networks (CNNs) have been widely adopted for diverse
computer vision tasks, including image classification~\cite{han2018new}, object
detection~\cite{ren2015faster}, anomaly detection~\cite{kim2018encoding} and
segmentation~\cite{kamnitsas2017efficient}. These models have demonstrated
impressive performance. 
Convolutional models have proven successful in image classification and object
recognition, as demonstrated in recent studies such as ResNet~\cite{he2016deep},
Very Deep Convolutional Networks~\cite{simonyan2014very}, and Understanding
Convolutional Networks~\cite{zeiler2014visualizing}.
These studies show that by increasing the number of convolutional layers, the
model becomes more robust in the feature extraction process~\cite{simonyan2014very}, however, deeper models present numerical difficulties
during training~\cite{he2016deep, ioffe2015batch}. The non-linearity of the
network at each layer reduces the gradients and that could lead to a very slow
training~\cite{ioffe2015batch,
glorot2010understanding}. One solution for this is to use batch normalization~\cite{ioffe2015batch} as an intermediate layer for the networks, which helps
to stabilize training and reduce the number of training epochs. Another solution
is given by Residual neural network (ResNet)~\cite{he2016deep}, which can stack
even hundreds of layers skipping the non-linearity by passing information of
previous layers directly. It is worthy noting that ResNet also uses batch
normalization layers. 

In another line of work, the Gradient Boosting Machines (GBMs) decision tree
ensembles~\cite{friedman2001greedy, mason1999boosting, ke2017lightgbm, chen2015xgboost} have become the state of the art for solving tabular
classification and
regression tasks~\cite{shwartz2022tabular, bentejac2021comparative}. GBMs work
by training a sequence of regressor models that sequentially learn the {\it
information} not learnt by previous models. This is done by computing the
gradients of the training data with respect to the previous iteration and by
fitting the following model to those gradient values or pseudo-residuals. The
final model combines all generated models in an additive manner.
These ideas have also been applied to sequentially train Neural Networks
\cite{inproceedingsemami, huang2018learning, nitanda2018functional,
bengio2005convex}. 
In~\cite{nitanda2018functional}, a gradient boosting based approach that uses a
weight estimation model to classify image labels is proposed. The model is
designed to mimic the ResNet deep neural network architecture and uses boosting
functional gradient minimization~\cite{mason1999boosting}.
In addition, it also involves formulating linear classifiers and
feature extraction, where the feature extraction produces input for the linear
classifiers, and the resulting approximated values are stacked in a ResNet
layer.

In this paper, we propose a novel deep learning training architecture framework
based on gradient boosting, which includes two structures: GB-CNN and GB-DNN.
GB-CNN, or Gradient Boosted Convolutional Neural Network, is a CNN training
architecture based on convolutional layers. On the other hand, GB-DNN is a
simpler architecture based only on dense layers more specific for tabular data.
Both architectures explicitly use the gradient boosting procedure to build a set
of embedded NNs in depth. The approach adds one dense layer at each iteration to
a copy of the previous network. Then, the previous dense layers are frozen and
the weights of the new added dense layer are trained on the residuals of the
previous iteration. In the case of GB-CNN, also the previously trained
convolutional layers are fine-tuned at each boosting iteration. 
The weights of previous layers are frozen to simplify the training of the
network and to avoid overfitting to the residuals.


The rest of the paper is organized as follows: In Section II, we review related
works in the field. In Section III, we provide an overview of our proposed
approach. In Section IV, we describe our experimental setup and present the
results of our experiments in Section V, finally, we conclude the paper in the
last section.

\section{Related Works}

Several studies have used the ideas of gradient boosting optimization to build
Neural Networks. For instance, in~\cite{bengio2005convex} they propose a convex
optimization model for training a shallow Neural Network that could reach the
global optimum. This is done by adding one hidden neuron at a time to the
network, and re-optimizing the whole network by including a $L^1$ regularization
on the top layer. This top layer serves as a regularizer to effectively remove
neurons. However, the model is computationally feasible only for very small
number of inputs attributes. In fact, the method is tested experimentally only
for 2D datasets. 
In another line of work, a shallow neural network is sequentially trained as an
additive expansion using gradient boosting \cite{emami2019sequential}.
The weights of the trained models
are stored to form a final neural network. Their idea is to build a
single network using a sequential approach avoiding having an ensemble of
networks. The work is specific for tabular multi-output regression problems
whether the method proposed in this article is a deep architecture valid for
both tabular and structured datasets, such as image datasets.

Furthermore, a development of ResNet in the context of boosting theory was
proposed in~\cite{huang2018learning}. This model, called BoostResNet, uses
residual blocks 
that are trained during boosting iterations based on~\cite{freund1997decision}.
The BoostResNet builds an ensemble of shallow blocks. In a similar proposal,
a deep ResNet-like model (ResFGB) is developed in depth by using a linear
classifier and gradient-boosting loss minimization~\cite{nitanda2018functional}.
The proposed method is different from these studies in distinct
aspects. In contrast to the work presented in \cite{huang2018learning, nitanda2018functional}
the proposed method is based on gradient boosting \cite{friedman2001greedy}.
Contrary to~\cite{nitanda2018functional}, the underlying architecture of our
work can work with any standard deep architecture (with or without convolutional
layers) rather than an ad-hoc specific network block. 
When trained with convolutional layers, the proposed method includes a series of
dense layers trained sequentially while jointly fine-tuning previously fitted
convolutional layers. In addition, the proposed method reduces the non-linearity
complexity by freezing the previously trained dense layers. 
On the other hand, ResFGB uses a combination of a linear classifiers and a
feature extractor, which is updated by stacking a resnet-type layer at each
iteration.
The~\cite{huang2018learning} study built layer-by-layer a ResNet boosting
(BoostResNet) over features.
BoostResNet offers a computational advantage over the conventional ResNet due to its reduced complexity, even though the reported performance is not consistently superior.

\section{Methodology}
In this paper we propose a methodology to train a set of Deep Neural Network
models as an additive expansion trained on the residual of a given loss
function. The procedure works by adding a new dense layer sequentially to a copy
of the previously trained deep regression NN with all dense layers from previous iterations
frozen. The motivation behind freezing the trained dense layers is that the
model is less likely to adapt to the noise in the data and avoid overfitting.
Based on this idea two deep architectures are proposed: one using convolutional
layers (GB-CNN) and one with only dense layers (GB-DNN).
In the case of GB-CNN, at each iteration, the model learns from the errors made
by the previous dense layers also fine-tuning the parameters of the
convolutional layers. 
In GB-DNN architecture, only the newly inserted dense layer is trained at each
iteration while simultaneously freezing all previously trained dense layers. 
In this section, we described the backbone of the proposed methods alongside with
its mathematical framework, followed by the used structure for the convolutional
layers. 
In the first subsection, we defined the mathematical framework of the GB-CNN and GB-DNN,
followed by the description of the CNN structure used in the GB-CNN method. 

\subsection{Gradient Boosted Convolutional and Deep Neural Network}

For the GB-CNN architecture, we assume the training and test instances with $D =
{(\mathbf{X}_i, \mathbf{y}_i)}_1^N$ distribution on input $\textbf{X}_i \in
\mathbb{R}^{B \times {P_H} \times {P_W} \times {Ch}}$ a 4-dimensional tensor,
where the first dimension $\mathbb{R}^{B}$ corresponds to the batch size, the
second and third dimensions $\mathbb{R}^{{P_H} \times {P_W}}$ represent the
height and width of each image respectively, and the fourth dimension
$\mathbb{R}^{Ch}$ represents the number of color channels. The labels
$\mathbf{y}_i$ are characterized by a one-hot encoding vector $y_i = [y_{i,1},
y_{i,2}, \dots, y_{i,K}]$, where $K$ is the number of classes, and $y_{i,j} = 1$
if the $i^{th}$ data point belongs to class $j$, and $y_{i,j} = 0$ otherwise. For the GB-DNN model, we assume the training and test instances with $D = {(\mathbf{X}_i, \mathbf{y}_i)}_1^N$ distribution on input $X_i \in \mathbb{R}^Fe$ with $Fe$ features and Response variables with $K$ attributes $y_i \in [1, K]$ respectively.

The idea is to learn and update the trainable parameters of the model 
to accurately estimate the class of unseen data 
by minimizing the cross-entropy $\ell (\mathbf{y}, p)$ loss function
\begin{align}\label{eq:loss}
\ell(\mathbf{y}_i, \mathbf{P}_i) = - \sum_{k=0}^{K} y_{i,k} \log p_{i,k},
\end{align} 
where $\mathbf{P}=\{\{p_{i,k}\}_{k=1}^{K}\}_{i=1}^{N}$ is a probability
matrix where $p_{i,k}$ is the estimate of the probability for the $i$-{th}
instance of belonging to class $k$. These probabilities are obtained from the
raw outputs of the trained regression networks, $F_k$, by applying a softmax function to
the outputs of the additive model
\begin{equation}
p_{k}(\cdot)=\frac{\exp(F_k(\cdot))}{\sum_{l=1}^{K}\exp(F_l(\cdot))} .
\end{equation}
The final model is built as an additive model of the outputs by combining
several networks $S_t$ iteratively 
\begin{align}\label{eq:additive_form}
F_t(\mathbf{X}_i) = F_{t-1}(\mathbf{X}_i) + \mathbf{\rho_t} S_t(\mathbf{X}_i),
\end{align} 
where $\mathbf{\rho_t}$ is a vector of weights for each class of the $t-{th}$
additive model $S_t(\mathbf{X}_i)$.

This additive model is built in a step-wise manner. In the GB-CNN, first, a
series of layers, including convolution, activation, pooling, batch
normalization, dense and flattening layers, is defined as the function $C$
\begin{align}\label{eq:cnn}
C(\mathbf{X}_i; \mathbf{\Omega}) = H(\mathbf{X}_i),
\end{align}
where $\mathbf{\Omega}$ is the set of trainable parameters of the model that
include weights and biases of all layers, and the function $H(\mathbf{X}_i)$ represents the output of the network, which is obtained by applying the set of learnable parameters $\mathbf{\Omega}$ to the input $\mathbf{X}_i$.
While in GB-DNN a first dense layer is initialized with random weights. 
The model construction is followed by adding a new linear
transformation with an activation function, representing the $t$-th
dense layer (boosting iteration)
\begin{align}\label{eq:dense}
S_t(\mathbf{M}_{i}; \mathbf{W}_t) = ReLU_t(\mathbf{W}_t \mathbf{M}_{i} + \mathbf{b}_i),
\end{align} 
where $\mathbf{W}_t$ is the weight matrix and $\mathbf{b}_t$ is the bias
vector for the $t$-{th} dense layer respectively, $\mathbf{M}_{i}$ is the
feature mapped output from the previous dense layer output for $i$-{th} 
instance, and $ReLU(x)=max(0,x)$ is the activation
function for the $t$-{th} dense layer. 

Hence, we can define the additive model of the $t$-{th} boosting iteration to
train the parameters for the proposed model
\begin{align}\label{eq:additive_gbcnn}
S_t = S_t (S_{t-1} & \bigl(C(\mathbf{X}_i; \mathbf{\Omega}_t)); \mathbf{W}_t
\bigl).
\end{align}
by minimizing the loss function $\ell$ (Eq.\ref{eq:loss}), using the following objective function,
\begin{align}\label{eq:optimization}
\begin{split}
\bigl(\rho_t, S_t \bigl) = \argmin_{(\rho_t, S_t)} \sum_{i=1}^N \ell(y_i, F_{t-1} (\mathbf{X}_i) + \rho_t S_t ).
\end{split}
\end{align}
where the trainable parameters of $S_{t-1}$ were trained in the $(t-1)$-{th}
iteration and are now frozen. The optimization process is the same for GB-CNN
and GB-DNN although the GB-CNN fine-tunes the convolutional layer at each
iteration whether GB-DNN only trains the newly added layer.

The optimization problem (Eq. \ref{eq:optimization}) continues by training the
new model on the pseudo-residuals of the previous epoch, $r_{i, t-1}$, which
are parallel to the gradient of the loss for step $t-1$
\begin{align}\label{eq:residual}
r_{i, t-1} &= - \left. \frac{\partial \ell (y_i , F(\mathbf{X}_i))}{\partial
F(\mathbf{X}_i)}.
\right |_{F(\mathbf{X}_i)=F_{t-1}(\mathbf{X}_i)}.
\end{align}
This is done using a regression neural network optimized using the square loss.
Since the network is trained using a loss that is different from the desired
cross-entropy loss (Eq.~\ref{eq:loss}), we then update the weights of the output
layer for each $K$ class label. This is done by computing the $\mathbf{\rho}_t$
vector for the $K$ class labels with a line search optimization
\begin{align}\label{eq:line_search}
f(\rho_t) = \sum_{i=1}^{N} \sum_{k=1}^{K} y_{i,k} ln \left (\frac{exp(F_{t-1,k}(X_i) +
\rho_{t,k} S_{t,k}(X_i)}{\sum_j^K exp(F_{t-1,j}(X_i) + \rho_{t,j} S_{t,j}(X_i)} \right ),
\end{align}
With this line search, a value of $\mathbf{\rho}_k$ is obtained for each
class. These values are then integrated into the neural network model by
multiplying all the weight values from the last layer to the $k$-th output
neuron by $\rho_k$ for $k=1 \dots K$.
%

In addition, an additional regularization term is included in the proposed
models, namely shrinkage rate, which is a constant value $\nu \in (0, 1] $ that
reduces the contribution of each additive model in the training procedure by
$\nu$ in order to prevent overfitting~\cite{friedman2001greedy}.
The schematic description of the proposed GB-CNN method is shown in Algorithm.
\ref{algo:deep_gbnn}.

\begin{algorithm}
 \caption{Training procedure of GB-CNN.}
 \label{algo:deep_gbnn}
 \begin{algorithmic}[h!]
 \renewcommand{\algorithmicrequire}{\textbf{Input:}}
 \renewcommand{\algorithmicensure}{\textbf{Output:}}
 \REQUIRE Input image data and related labels $D=\{\mathbf{x}_i,y_i\}_{1}^N$
 \REQUIRE Number of boosting iterations $T$.
 \REQUIRE Training epoch on mini-batches $E$.
 \REQUIRE Gradient Boosting loss function $\ell$ (Eq.\ref{eq:loss}).
 \STATE Image batch generator $G(D) = \{(\mathbf{x}_i, y_i)\}_{1}^{N}$.
 \FOR {$e = 0$ to $E$}
 \STATE Fit the GB-CNN with one dense layer to images-residual.
 \IF {if the training additive loss converges}
  \STATE break
  \ENDIF 
 \ENDFOR
 \STATE Update the trainable parameters $\mathbf{W}_0$.
 \STATE Freeze the added dense layer's parameters $\mathbf{W}_0$.
  \FOR {$t = 1$ to $T-1$}
  \STATE Add a new dense layer.
  \FOR {$e = 0$ to $E$}
  \STATE Fit the GB-CNN to images-residual.
  \IF {if the training additive loss converges}
  \STATE break
  \ENDIF
  \ENDFOR
  \STATE Update the trainable parameters $\mathbf{W}_t$.
  \STATE Freeze the added dense layer's parameters $\mathbf{W}_t$.
  \IF {if the training Gradient Boosting loss converges}
  \STATE break
  \ENDIF
  \ENDFOR
  \STATE Update the CNNs' weights.
 \RETURN A fully trained and fine-tuned GB-CNN network.
 \end{algorithmic} 
 \end{algorithm}

\subsection{Convolutional layers design}

Despite the fact that the proposed method could be applied to different CNN
architectures, in the following, we describe the design of convolutional and
dense layers for the model applied in the experiments.

The applied CNN architecture consists of a sequence of three blocks. Each block
has two 2D-convolutional layers, a batch normalization, max pooling and a
dropout layer. The dense layers are then connected to the last of these three
layers. The configuration of the blocks is the following. The two
2D-convolutional layers in the first block consists of $32 \times 3 \times 3$
filters and {\it ReLU} as the activation function. The subsequent block 
has two more 2D-convolutional layers with 64 filters. The
convolutional layers of the final block are of size 128.
The concatenated layers are two 2D-convolutional layers with 128 filters. All
2D-convolutional layers use filters of size (3x3). 
After the second convolutional layer of each block, batch normalization is
applied to stabilize the distribution of activations. Then a max pooling layer
of size (2, 2) is applied to the output to reduce the spatial dependence of the
feature maps and increase the invariance to small translations, and finally a
dropout layer is included to prevent overfitting with 0.2, 0.3 and 0.4 for each
block respectively. Finally, the output is flatten and connected to fully
connected dense layers. 
The dense layers are added and trained iteratively while freezing previous dense
layers. This model is compared to the same architecture trained jointly as a
standard CNN model.
The second proposed architecture considers only the iterative training of a
network composed only of dense layers.

\begin{figure}[htbp]
\centering
\includegraphics[width=8cm]{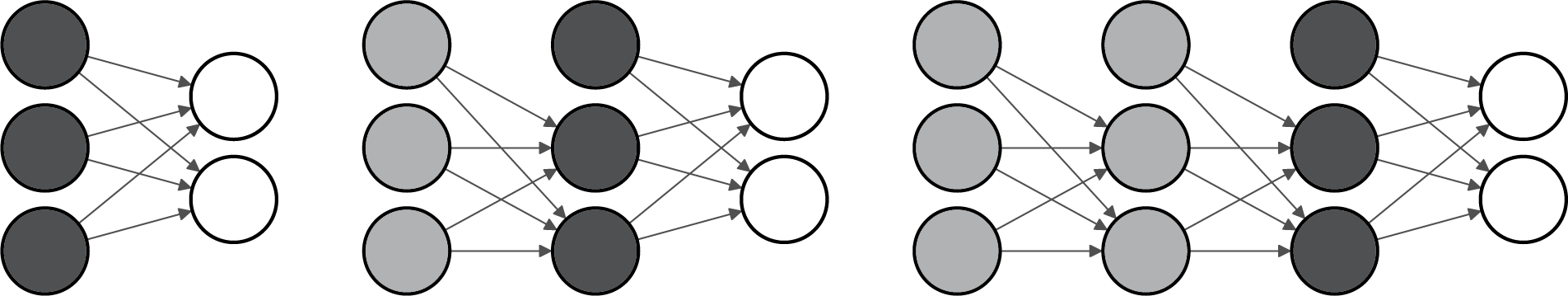}
\caption{A snippet of the iterative model training procedure. Light-gray indicates that the layer is frozen, dark gray that the layer
is being trained and white indicates output layer (also being trained)}
\label{fig1:dense}
\end{figure}

This process is illustrated in Fig.~\ref{fig1:dense} in which only the dense
layers are shown (Note, that the figure is not representing the actual size of
the used dense layers). In the first, iteration, one single dense (and output)
layer is trained (dark grey units in left diagram of Fig.~\ref{fig1:dense}).
This layer is the first fully connected layer. After fitting this model,
the model is copied and a second dense layer is added (iteration 1 and second
diagram in Fig.~\ref{fig1:dense})
freezing the weights of the first dense layer froze (shown in the
Fig.~\ref{fig1:dense} with light gray neurons). This second step fine-tunes the
parameters of the convolutional layers (if present), skips the training of the
previous dense layer, and trains the newly added dense layer (dark gray units).
Each new model, fits the last dense layers and the convolutional blocks to the
corresponding pseudo-residuals. The training procedure continues until
convergence. 

We carried out several preliminary analysis in which we compared the performance
of the presented method with respect to the same procedure but without applying
the freezing of previous layers. These experiments showed that freezing improved
the performance. This could be explained as new models are trained on smaller
residuals and including more complex models (i.e. unfrozen) could lead to
overfitting.

\section{Experiments}

In order to evaluate the proposed new methods structure for Convolutional
(GB-CNN) and Deep Neural Networks (GB-DNN), several supervised 2D-image
classification tasks and tabular datasets were considered. In the image
datasets, the proposed model (GB-CNN) is compared with respect to a CNN model
that uses the same architecture and configuration, although with more dense
layers (more details below).
In the tabular datasets, GB-DNN is compared with a deep neural network with same
settings and layers.
The objective of our experiments was to
evaluate the performance of the tested methods in
terms of accuracy and to determine its usefulness for solving the 2D-image and
tabular classification problems. The code of the proposed models is made
available on github as
GB-CNN\footnote{\href{https://github.com/GAA-UAM/GB-CNN}{github.com/GAA-UAM/GB-CNN}}.



\begin{table}[h!]
  \centering
  \fontsize{9}{9}\selectfont 
  \caption{Dataset Properties for Image Classification Tasks}

  \begin{tabular}{p{3.0cm} p{1.6cm} p{1.25cm} p{0.3cm} p{0.3cm}}
	\hline
    \textbf{Name}  & \textbf{train/test} & \textbf{$P_H \times P_W$} & \textbf{$K$} &  \textbf{$Ch$} \\
    \hline
    MNIST \cite{mu2019mnist} & 60,000/10,000  & $28 \times 28$ & 10 & $1$ \\ 
    CIFAR-10 \cite{Krizhevsky09learningmultiple} &  50,000/10,000 & $32 \times 32$ & 10 & $3$ \\ 
    Rice varieties \cite{koklu2021classification} & 56,250/18,750 & $32 \times 32$ & 5 & $3$ \\
    Fashion-MNIST \cite{xiao2017fashion} & 60,000/10,000 & $28 \times 28$ & 10 & $1$ \\ 	
    Kuzushiji-MNIST \cite{clanuwat2018deep} & 60,000/10,000 & $28 \times 28$ & 10 & $1$ \\
    MNIST-Corrupted \cite{clanuwat2018deep} & 60,000/10,000 & $28 \times 28$ & 10 & $1$ \\
    Rock-Paper-Scissors \cite{rps} & 2,520/370 & $32 \times 32$ & 3 & 3 \\ 
    \hline
  \end{tabular}
  \label{tab:dataset_image}
\end{table}

Regarding the 2D-image classification problem, seven 2D-image datasets of
various areas of application, class labels, instances, pixel resolution, and
color channels are used in this study, as described in
Table~\ref{tab:dataset_image} for the convolutional models. In this experiment,
a data generator was used to generate
batches of training data on-the-fly during the training process. This allowed us
to train the model on a large dataset, without having to load the entire dataset
into GPU memory. Moreover, the generator shuffled the data, applied data
augmentation techniques (including rescaling the pixel values), and yielded
batches with a size of 128.  The common hyperparameters and settings of the used
GB-CNN and CNN were the same, including the structure of
the convolutional layers (described before), the number of hidden neurons in
each dense (set to 20), a learning rate of 0.001, and 200 epochs for training
with early stopping to prevent overfitting.
The shrinkage rate of GB-CNN, used to combine the different
generated CNNs, was set to 0.1. As previously described, the proposed method
adds one dense layer at each iteration to a copy of the model of the previous
iteration. The maximum number of dense
layers/iterations was set to 10. Although, GB-CNN generally converges after 2 or
3 iterations. This will be shown in the first experiment in which we analyze the
evolution of the loss and accuracy for the tested models. The
number of dense layers for CNN was left at 10. 
Furthermore, we conducted an additional experiment wherein three images ({\it
MNIST}, {\it Fashion-MNIST}, and {\it CIFAR-10}) were analyzed using two dense
layers, each with a size of 128 and identical settings, in order to investigate
the impact of the size of the dense layers in the experimental outcomes.

\begin{table}[h!]
  \centering
  \fontsize{9}{9}\selectfont 
  \caption{Dataset Properties for Image Classification Tasks}
  \begin{tabular}{p{2.0cm} p{1.85cm} p{1.25cm} p{1.5cm}}
	\hline
     \textbf{Name}  & \textbf{instances} &  \textbf{Features} & \textbf{class labels} \\
    \hline
    Digits~\cite{UCI} &  1797 &  64 & 10 \\ 
    Ionosphere~\cite{UCI}  & 351  & 34 & 2 \\ 
    Letter-26~\cite{UCI}  & 20,000  & 16 & 26 \\ 
    Sonar  & 208 & 2 & 60 \\ 
    USPS~\cite{hull1994database}  & 9,298 & 256 & 10 \\ 	
    Vowel~\cite{UCI}  & 990 & 10 & 11 \\
    Waveform~\cite{UCI} & 5,000 &  21 & 3 \\
    \hline   
  \end{tabular}
  \label{tab:dataset_tabular}
\end{table}

Moreover, in this work, tabular classification datasets from different sources
\cite{UCI,hull1994database} were also tested. The characteristics of these
datasets are shown in Table~\ref{tab:dataset_tabular}.
The classifications tasks are of different types, ranging from radar
data to handwritten digits and vowel pronunciation. The hyperparameter values of
the proposed GB-DNN and the standard Deep-NN are identical. For these datasets,
the training was performed using 10-fold cross-validation. In order to determine
the optimal values for hyper-parameter of the models, a with-in train grid search
process was applied. The test values for the hyper-parameters are: learning rate
of both models (ranging from 0.1 to 0.001) and shrinkage rate $\nu$ (ranging
from 0.1 to 1.0) of the GB-DNN model. Both models are composed of three dense
layers, each with a size of 100 neurons and ReLU as the activation function.

\section{Results}

\begin{figure}[t]
\centering

\includegraphics[width=8cm]{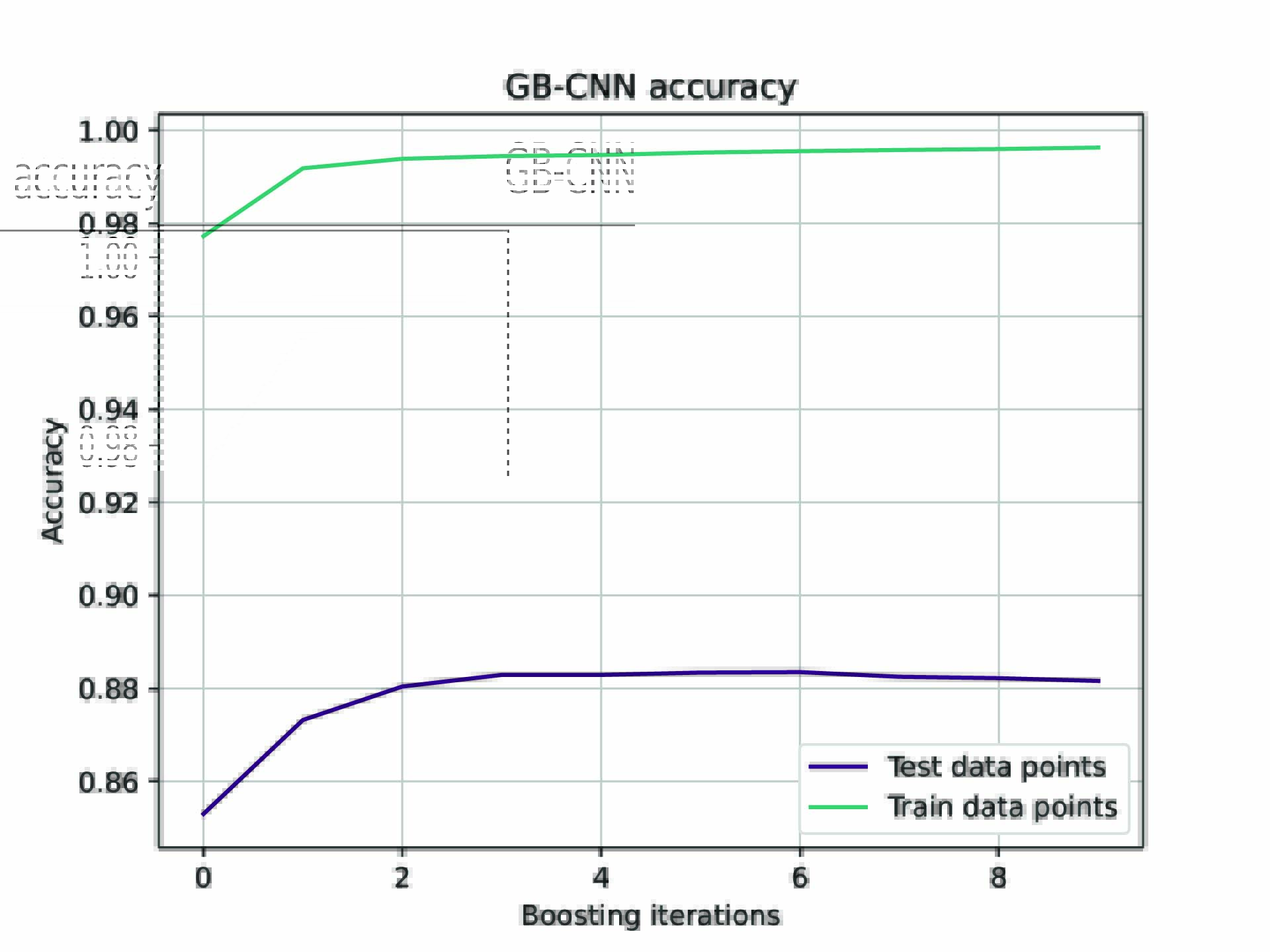}
\caption{GB-CNN train and test accuracy on CIFAR10 with respect to the
number of boosting iterations}
\label{fig:gbloss1}
\end{figure}

\begin{figure}[t]
\centering

\includegraphics[width=8cm]{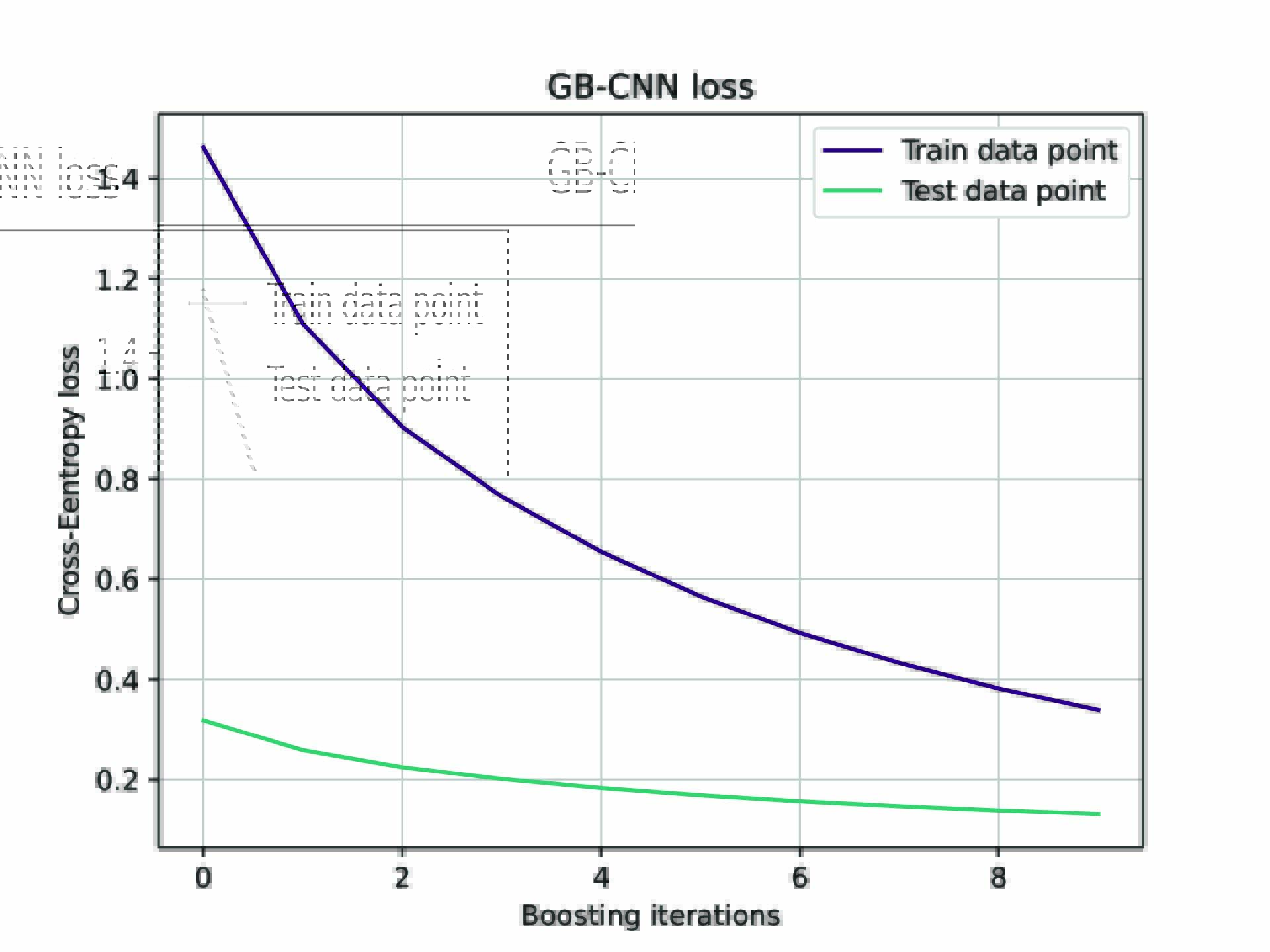}
\caption{GB-CNN train and test cross-entropy loss on CIFAR10 with respect to the number of
boosting iterations}
\label{fig:gbloss2}
\end{figure}

\begin{figure}[t]
\centering
\includegraphics[width=8cm]{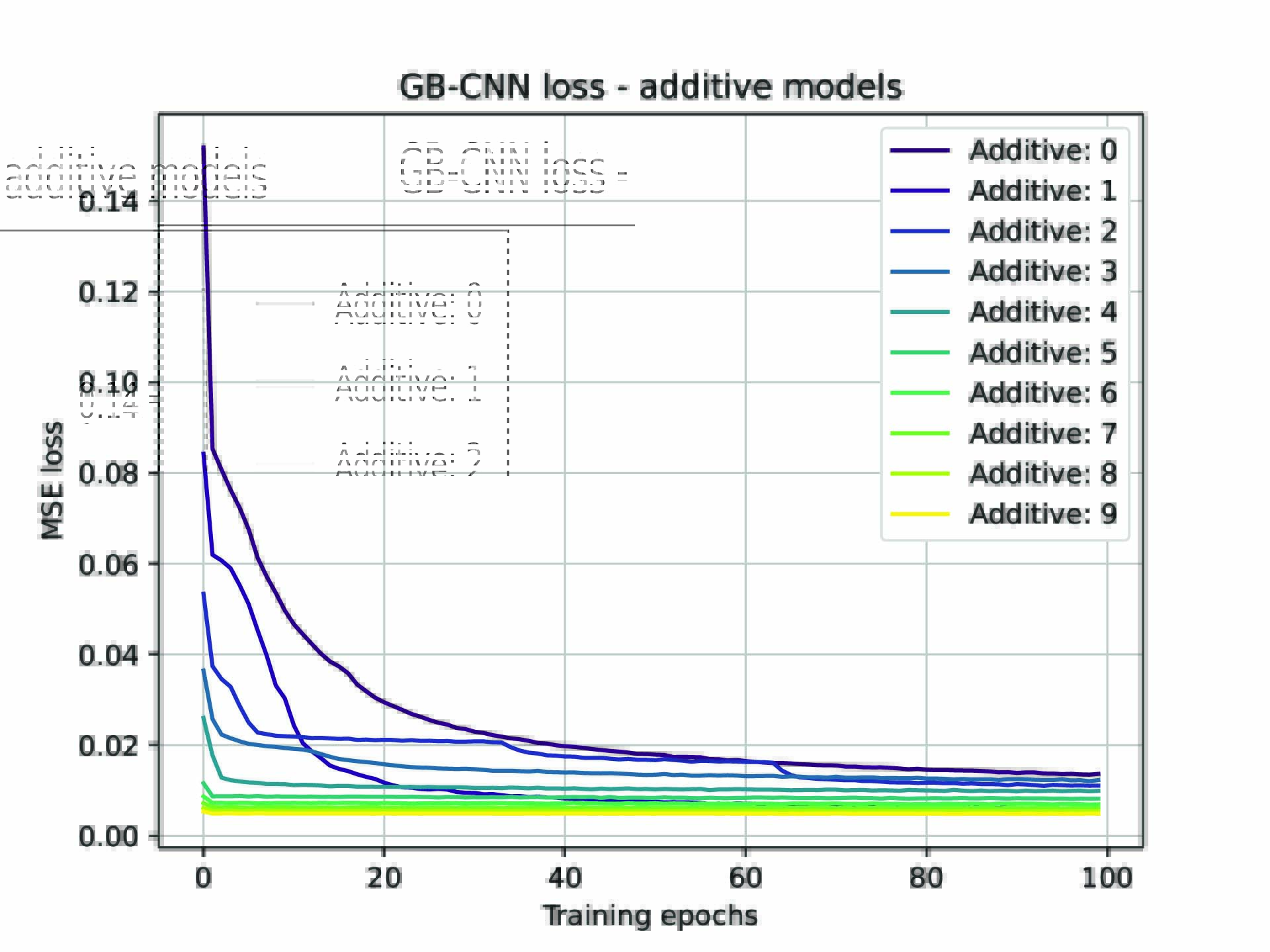}
\caption{GB-CNN - Additive model train MSE loss on CIFAR10 with respect to the number of
boosting iterations}
\label{fig:gbloss3}
\end{figure}

\begin{figure}[t]
\centering
\includegraphics[width=8cm]{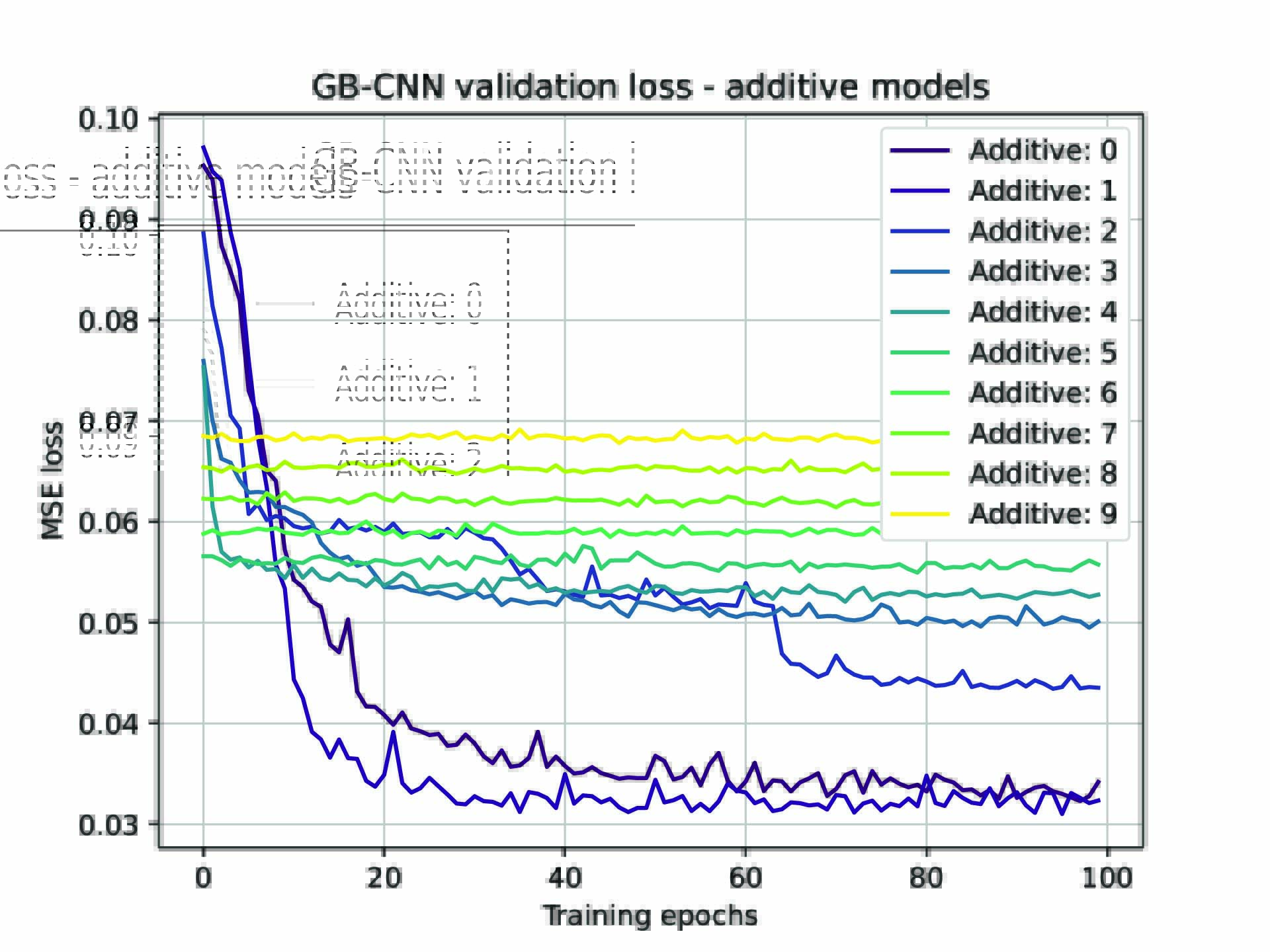}
\caption{GB-CNN - Additive model test MSE loss on CIFAR10 with respect to the number of
boosting iterations}
\label{fig:gbloss4}
\end{figure}

\begin{figure}[t]
\centering
\includegraphics[width=8cm]{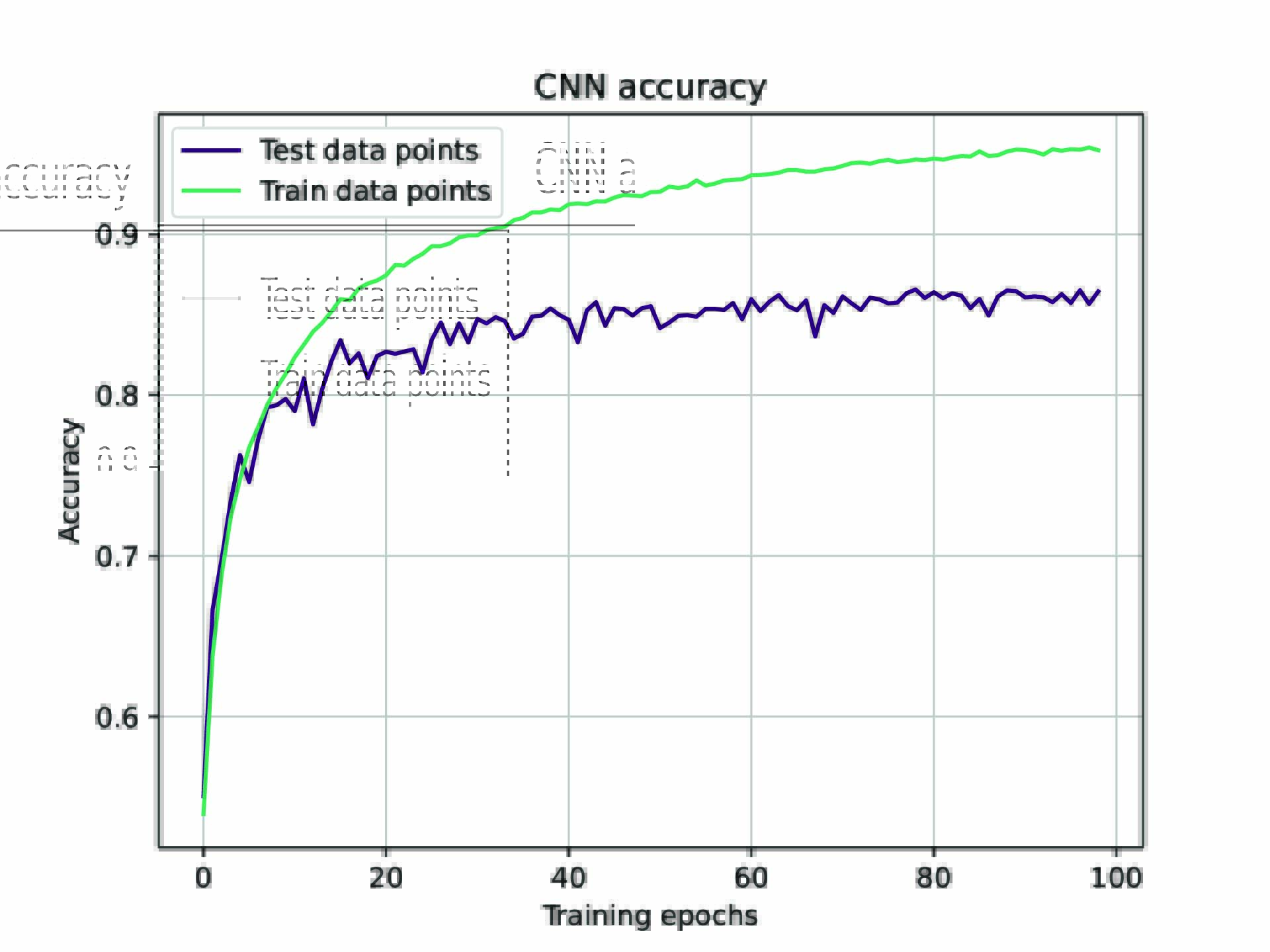}
\caption{CNN train and test accuracy on CIFAR10 with respect to the training epochs}
\label{fig:cnnloss1}
\end{figure}

\begin{figure}[t]
\centering
\includegraphics[width=8cm]{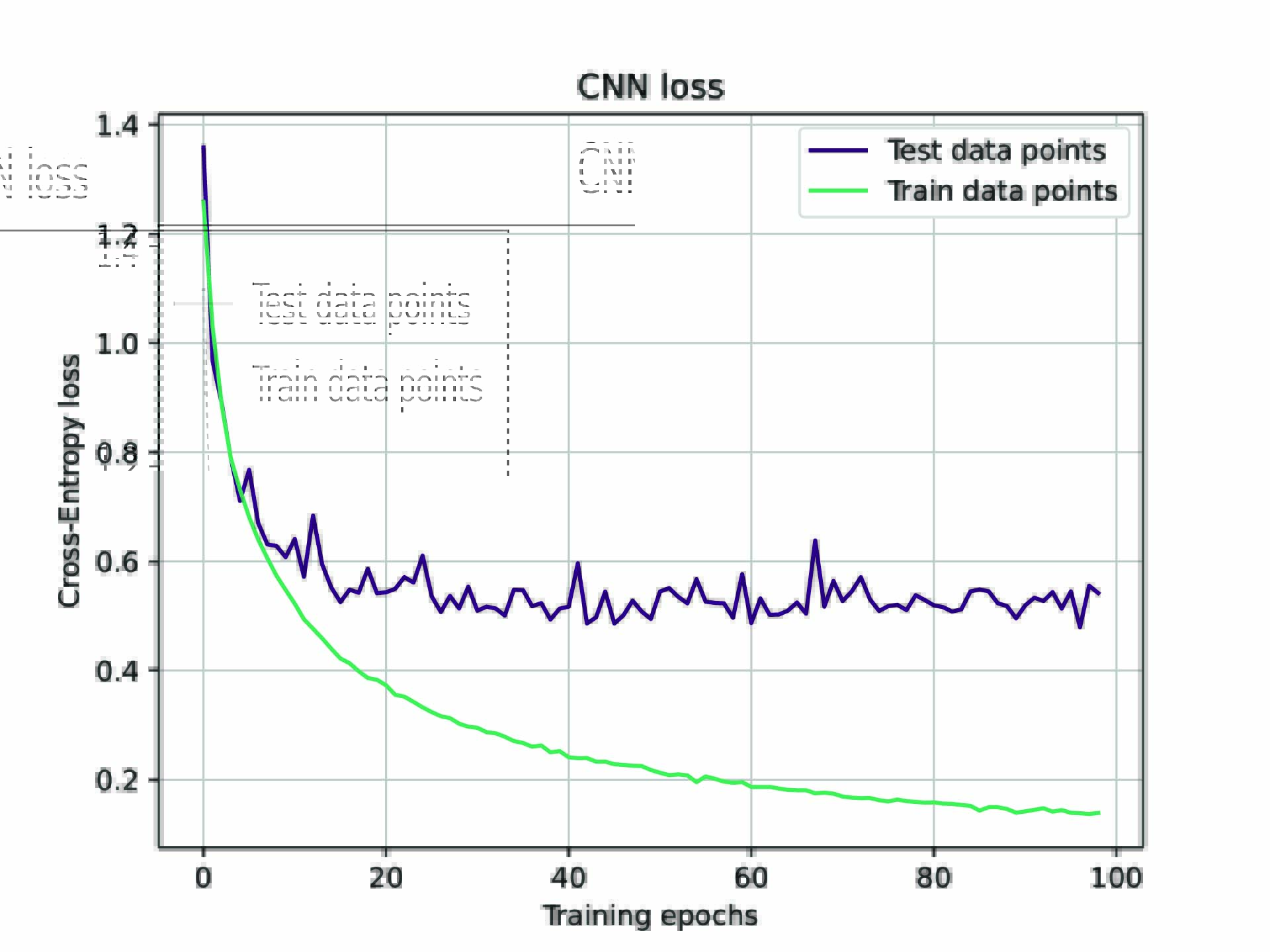}
\caption{CNN train and test cross-entropy loss on CIFAR10 with respect to the training epochs}
\label{fig:cnnloss2}
\end{figure}

In order to analyze the convergence rate of the proposed method, a first batch
of experiments was carried out that registered the loss and accuracy of the
method after each iteration.
We further monitored the loss
function across various additive models during training for GB-CNN. The
experiment was conducted using the {\it CIFAR-10} dataset, and the evolution of
the loss and accuracy was measured in both training and test data points. Also
the standard CNN was monitored in the same way.
Both models were configured with the same hyperparameters,
including identical convolutional layers as outlined in Section III subsection B, and 10
dense layers with a size of 20. The learning rate was set to $0.001$, with a
batch size of $128$, and $100$ training epochs for both models. In the case of
GB-CNN, the shrinkage was set to $0.01$. 
The results are shown in Figures~\ref{fig:gbloss1}--\ref{fig:gbloss4}, and
Figures~\ref{fig:cnnloss1} and \ref{fig:cnnloss2}). 

The evolution performance for GB-CNN is shown in
Figs.~\ref{fig:gbloss1}--\ref{fig:gbloss4}.
Fig.~\ref{fig:gbloss1} illustrates the train and test accuracy of the model
with respect to the number of boosting iterations. In Fig.~\ref{fig:gbloss2},
the average cross-entropy loss across various boosting iterations is shown.
Additionally, Figs.~\ref{fig:gbloss3} and \ref{fig:gbloss4} provide an
overview of the evolution of the mean square error (MSE) of the additive models
created during the boosting iterations with respect to the number of epochs.
Note that the loss here is measured as MSE since the individual models trained during boosting are regression networks.
From plots in Figs.~\ref{fig:gbloss1} and \ref{fig:gbloss2}, it can be observed
that the proposed model converges after
few booting iterations. Specifically, after three boosting iterations the model
has converged and the training process could be stopped. 
A similar tendency can be observed from the learning evolution of the
individual trained networks shown in Figs.~\ref{fig:gbloss3} for train and
\ref{fig:gbloss4} for test. We can observe that the first models reduce the loss
more significantly. Again, after a few iterations the benefit of adding more
models is reduced. Based on these results, for the next experiments we will
limit the number of boosting iterations to 2-3.

On the other hand, Figs.~\ref{fig:cnnloss1} and \ref{fig:cnnloss2} present the
performance of the CNN model with respect to the training epochs. Specifically,
the train and test accuracy and cross-entropy loss of the model are depicted in
Figs.~\ref{fig:cnnloss1} and Figs.~\ref{fig:cnnloss2}, respectively. 
As it can be observed from these plots, the CNN model has converged in
terms of test accuracy and loss after 100 epochs. This final performance is
slightly better to the one achieved by GB-CNN after the first iteration (see
Figs.~\ref{fig:gbloss1} and \ref{fig:gbloss2}). This makes sense since the first
boosting iteration is also a CNN model, although simpler since it has only one
dense layer. The most interesting point is that when more layers are included
(and models), the proposed GB-CNN architecture manages to further
learn at the point where the CNN has converged. In this way, the final
performance of the proposed method is better than that of the CNN. In the
following a more comprehensive comparison is carried out including the described
datasets.

Tables~\ref{tab:results_conv_1}, \ref{tab:results_conv_2} and
\ref{tab:results_tabular} show the generalization accuracy results estimated in
the test subsets for the analyzed datasets and models.
Table~\ref{tab:results_conv_1} show the results for the convolutional models
(GB-CNN and CNN) using 20 neuron in the dense hidden layers.
Table~\ref{tab:results_conv_2}
include the results for selected image datasets and with larger dense layers of
128 neurons. Finally, Table~\ref{tab:results_tabular} portray the results for
the deep models (GB-DNN and DNN) in the tabular datasets.
The best results of the tables are highlighted using a gray background.

As it can be observed in Table~\ref{tab:results_conv_1}, the generalization
accuracy of the proposed convolutional model (GB-CNN) 
is higher than that of the baseline CNN model across all tested datasets. 
The difference is specially large on the {\it Rock-Paper-Scissors}. The
proposed model achieved an accuracy of $87.37\%$, which is $19.36\%$ higher than
the accuracy of CNN ($68.01\%$). On the other datasets the differences are small
but consistently in favor of the proposed architecture. For instance, in {\it
CIFAR-10}, the proposed model achieved an accuracy of $87.65\%$ with respect to
the $86.71\%$ that achieves the CNN.
In the {\it MNIST} dataset, well-known for its simplicity and high accuracy that
the CNN models obtain, GB-CNN was able to achieve a remarkable accuracy of
$99.61\%$, which is higher than the accuracy of CNN by a very small margin
($0.06$).

\begin{table}[h]
  \centering
  \caption{Accuracy of GB-CNN and CNN models on the experiment with a 20-neuron dense layer. The best results are
  highlighted in gray}
  
  \begin{tabularx}{\columnwidth}{lXX}
	\hline    
    \textbf{2D-Image dataset}  & \textbf{GB-CNN} & \textbf{CNN}  \\
    \hline
    MNIST & \cellcolor{gray}{$99.61\%$} & $99.55\%$ \\ 
    CIFAR-10  & \cellcolor{gray}{$87.65\%$} & $86.71\%$\\ 
    Rice varieties  & \cellcolor{gray}{$99.82\%$} & $98.37\%$\\ 
    Fashion-MNIST & \cellcolor{gray}{$94.34\%$} & $93.61\%$ \\ 
    Kuzushiji-MNIST & \cellcolor{gray}{$98.40\%$} & $96.70\%$  \\
    MNIST-Corrupted & \cellcolor{gray}{$99.58\%$} & $99.35\%$  \\
    Rock-Paper-Scissors & \cellcolor{gray}{$87.37\%$} & $68.01\%$ \\ 
    \hline
  \end{tabularx}
  \label{tab:results_conv_1}
\end{table} 

In relation to the experiment involving the utilization of a dense layer with a
larger size (128), the accuracy results for the GB-CNN and CNN models are
presented in Table~\ref{tab:results_conv_2}. These models were trained and
evaluated on three distinct datasets: {\it MNIST}, {\it CIFAR-10}, and {\it Fashion-MNIST}. The
findings demonstrate that the GB-CNN model outperformed the CNN model across all
these three datasets. Nonetheless, closer examination of the table reveals that the results obtained
using 128 dense layers (as shown in Table~\ref{tab:results_conv_2}) are slightly
inferior to those using 20 neuron layers (as shown in
Table~\ref{tab:results_conv_1}). This implies that for these particular
datasets, employing a greater number of smaller layers may be more effective for
capturing the information of the tasks.


\begin{table}[h]
  \centering
  \caption{Accuracy of GB-CNN and CNN models on the experiment with a 128-neuron dense layer. The best results are
  highlighted in gray}
  
  \begin{tabularx}{\columnwidth}{lXX}
	\hline    
    2D-Image dataset  & GB-CNN & CNN  \\
    \hline
    MNIST & \cellcolor{gray}{$99.66\%$} & $99.40\%$ \\ 
    CIFAR-10  & \cellcolor{gray}{$86.72\%$} & $86.30\%$\\ 
    Fashion-MNIST & \cellcolor{gray}{$94.02\%$} & $92.38\%$ \\ 
    \hline
  \end{tabularx}
  \label{tab:results_conv_2}
\end{table}

The results for the tabular datasets summarized in
Table~\ref{tab:results_tabular} for GB-DNN and Deep-NN.
The results demonstrate that GB-DNN outperforms Deep-NN in most of the datasets.
In particular, GB-DNN achieved the highest accuracy in the {\it Digits}, {\it
Ionosphere}, {\it Letter-26}, {\it USPS}, {\it Vowel}, and {\it Waveform}
datasets. Deep-NN, on the other hand, achieved the highest accuracy in the Sonar
dataset. As in previous results, the differences are in general small with
higher difference in {\it Waveform} in favor of GB-CNN ($2.82$-point difference)
and in {\it Sonar} in favor of Deep-NN ($1.45$-point difference).
Finally, comparing the results of the proposed method with respect to the work
of~\cite{nitanda2018functional}, it can be observed that in the three common
tasks our method outperforms ResFGB in two tasks ({\it MNIST} and {\it USPS})
and is outperformed in one ({\it Letter-26}) with the advantage of our method of
being much simpler and composed of standard network layers. 

\begin{table}[h]
  \centering
  \fontsize{9}{9}\selectfont
  \caption{Accuracy of GB-DNN and Deep-NN models. The best results are
  highlighted in gray}
  
  \begin{tabularx}{\columnwidth}{lXX}
	\hline    
    \textbf{Tabular dataset}  & \textbf{GB-DNN} & \textbf{Deep-NN}  \\
    \hline
    Digits & \cellcolor{gray}{$98.22\%$} $\pm$1.24 & $97.55\%$ $\pm$1.06 \\ 
    Ionosphere & \cellcolor{gray}{$94.87\%$} $\pm$3.08 & $94.60\%$ $\pm$3.66 \\
    Letter-26 & \cellcolor{gray}{$95.50\%$} $\pm$0.49& $95.29\%$ $\pm$0.63 \\
    Sonar  & {$86.55\%$} $\pm$9.03 & \cellcolor{gray}{$88.00\%$} $\pm$3.78\\ 
    USPS  & \cellcolor{gray}{$97.39\%$} $\pm$0.40& {$96.95\%$} $\pm$0.62\\ 
    Vowel & \cellcolor{gray}{$98.08\%$} $\pm$1.59& $97.47\%$ $\pm$2.08\\
    Waveform & \cellcolor{gray}{$85.20\%$} $\pm$1.89& $82.38\%$ $\pm$2.37\\  

    \hline
  \end{tabularx}
  \label{tab:results_tabular}
\end{table} 


\section{Conclusion}


This paper presents a novel training approach for training a set of deep neural
networks based on convolutional (CNNs) and deep (DNNs) architectures, called
GB-CNN and GB-DNN respectively. The proposed training procedures are based on
gradient boosting algorithm, which trains iteratively a set of models in order
to learn the information not captured in previous iterations.
Additionally, both proposed models employ an inner dense layer freezing approach
to reduce model complexity and non-linearity.

To evaluate the effectiveness of the proposed models, we conducted experiments
on various 2D-images and tabular datasets, ranging from radar data to
handwritten digits, fashion, and agriculture areas. Our results demonstrate that
the proposed GB-CNN model outperforms traditional CNN models using the same
architecture in terms of accuracy in all the analyzed image datasets. 
Moreover, the GB-DNN model outperforms DNN models in terms of accuracy on most
of the studied tabular datasets. 

An analysis of the convergence of the proposed model showed that it converges
after few boosting iterations. More interestingly, during these iterations the
proposed methodology is able to improve the performance of the previous models
after the point in which a standard single CNN has converged.
This shows that the proposed GB-CNN and GB-DNN models represent a robust and
effective solution for 2D-image and tabular classification tasks. 
Further research and analysis of these models need to be carried out for
practical applications in the future.

\newpage
\bibliographystyle{unsrt}  
\bibliography{references}

\begin{thebibliography}{10}

\bibitem{NIPS2012_4824}
Alex Krizhevsky, Ilya Sutskever, and Geoffrey~E. Hinton.
\newblock Imagenet classification with deep convolutional neural networks.
\newblock In F.~Pereira, C.~J.~C. Burges, L.~Bottou, and K.~Q. Weinberger,
  editors, {\em Advances in Neural Information Processing Systems 25}, pages
  1097--1105. Curran Associates, Inc., 2012.

\bibitem{han2018new}
Dongmei Han, Qigang Liu, and Weiguo Fan.
\newblock A new image classification method using cnn transfer learning and web
  data augmentation.
\newblock {\em Expert Systems with Applications}, 95:43--56, 2018.

\bibitem{ren2015faster}
Shaoqing Ren, Kaiming He, Ross Girshick, and Jian Sun.
\newblock Faster r-cnn: Towards real-time object detection with region proposal
  networks.
\newblock {\em Advances in neural information processing systems}, 28, 2015.

\bibitem{kim2018encoding}
Taejoon Kim, Sang~C Suh, Hyunjoo Kim, Jonghyun Kim, and Jinoh Kim.
\newblock An encoding technique for cnn-based network anomaly detection.
\newblock In {\em 2018 IEEE International Conference on Big Data (Big Data)},
  pages 2960--2965. IEEE, 2018.

\bibitem{kamnitsas2017efficient}
Konstantinos Kamnitsas, Christian Ledig, Virginia~FJ Newcombe, Joanna~P
  Simpson, Andrew~D Kane, David~K Menon, Daniel Rueckert, and Ben Glocker.
\newblock Efficient multi-scale 3d cnn with fully connected crf for accurate
  brain lesion segmentation.
\newblock {\em Medical image analysis}, 36:61--78, 2017.

\bibitem{he2016deep}
Kaiming He, Xiangyu Zhang, Shaoqing Ren, and Jian Sun.
\newblock Deep residual learning for image recognition.
\newblock In {\em Proceedings of the IEEE conference on computer vision and
  pattern recognition}, pages 770--778, 2016.

\bibitem{simonyan2014very}
Karen Simonyan and Andrew Zisserman.
\newblock Very deep convolutional networks for large-scale image recognition.
\newblock {\em arXiv preprint arXiv:1409.1556}, 2014.

\bibitem{zeiler2014visualizing}
Matthew~D Zeiler and Rob Fergus.
\newblock Visualizing and understanding convolutional networks.
\newblock In {\em European conference on computer vision}, pages 818--833.
  Springer, 2014.

\bibitem{ioffe2015batch}
Sergey Ioffe and Christian Szegedy.
\newblock Batch normalization: Accelerating deep network training by reducing
  internal covariate shift.
\newblock In {\em International conference on machine learning}, pages
  448--456. PMLR, 2015.

\bibitem{glorot2010understanding}
Xavier Glorot and Yoshua Bengio.
\newblock Understanding the difficulty of training deep feedforward neural
  networks.
\newblock In {\em Proceedings of the thirteenth international conference on
  artificial intelligence and statistics}, pages 249--256. JMLR Workshop and
  Conference Proceedings, 2010.

\bibitem{friedman2001greedy}
Jerome~H Friedman.
\newblock Greedy function approximation: a gradient boosting machine.
\newblock {\em Annals of statistics}, pages 1189--1232, 2001.

\bibitem{mason1999boosting}
Llew Mason, Jonathan Baxter, Peter Bartlett, and Marcus Frean.
\newblock Boosting algorithms as gradient descent.
\newblock {\em Advances in neural information processing systems}, 12, 1999.

\bibitem{ke2017lightgbm}
Guolin Ke, Qi~Meng, Thomas Finley, Taifeng Wang, Wei Chen, Weidong Ma, Qiwei
  Ye, and Tie-Yan Liu.
\newblock Lightgbm: A highly efficient gradient boosting decision tree.
\newblock {\em Advances in neural information processing systems}, 30, 2017.

\bibitem{chen2015xgboost}
Tianqi Chen, Tong He, Michael Benesty, Vadim Khotilovich, Yuan Tang, Hyunsu
  Cho, Kailong Chen, et~al.
\newblock Xgboost: extreme gradient boosting.
\newblock {\em R package version 0.4-2}, 1(4):1--4, 2015.

\bibitem{shwartz2022tabular}
Ravid Shwartz-Ziv and Amitai Armon.
\newblock Tabular data: Deep learning is not all you need.
\newblock {\em Information Fusion}, 81:84--90, 2022.

\bibitem{bentejac2021comparative}
Candice Bent{\'e}jac, Anna Cs{\"o}rg{\H{o}}, and Gonzalo
  Mart{\'\i}nez-Mu{\~n}oz.
\newblock A comparative analysis of gradient boosting algorithms.
\newblock {\em Artificial Intelligence Review}, 54:1937--1967, 2021.

\bibitem{inproceedingsemami}
Seyedsaman Emami and Gonzalo Martínez-Muñoz.
\newblock Multioutput regression neural network training via gradient boosting.
\newblock pages 145--150, 01 2022.

\bibitem{huang2018learning}
Furong Huang, Jordan Ash, John Langford, and Robert Schapire.
\newblock Learning deep resnet blocks sequentially using boosting theory.
\newblock In {\em International Conference on Machine Learning}, pages
  2058--2067. PMLR, 2018.

\bibitem{nitanda2018functional}
Atsushi Nitanda and Taiji Suzuki.
\newblock Functional gradient boosting based on residual network perception.
\newblock In {\em International Conference on Machine Learning}, pages
  3819--3828. PMLR, 2018.

\bibitem{bengio2005convex}
Yoshua Bengio, Nicolas Roux, Pascal Vincent, Olivier Delalleau, and Patrice
  Marcotte.
\newblock Convex neural networks.
\newblock {\em Advances in neural information processing systems}, 18, 2005.

\bibitem{emami2019sequential}
Seyedsaman Emami and Gonzalo Mart{\'\i}nez-Mu{\~n}oz.
\newblock Sequential training of neural networks with gradient boosting.
\newblock {\em arXiv preprint arXiv:1909.12098}, 2019.

\bibitem{freund1997decision}
Yoav Freund and Robert~E Schapire.
\newblock A decision-theoretic generalization of on-line learning and an
  application to boosting.
\newblock {\em Journal of computer and system sciences}, 55(1):119--139, 1997.

\bibitem{mu2019mnist}
Norman Mu and Justin Gilmer.
\newblock Mnist-c: A robustness benchmark for computer vision.
\newblock {\em arXiv preprint arXiv:1906.02337}, 2019.

\bibitem{Krizhevsky09learningmultiple}
Alex Krizhevsky.
\newblock Learning multiple layers of features from tiny images.
\newblock Technical report, 2009.

\bibitem{koklu2021classification}
Murat Koklu, Ilkay Cinar, and Yavuz~Selim Taspinar.
\newblock Classification of rice varieties with deep learning methods.
\newblock {\em Computers and electronics in agriculture}, 187:106285, 2021.

\bibitem{xiao2017fashion}
Han Xiao, Kashif Rasul, and Roland Vollgraf.
\newblock Fashion-mnist: a novel image dataset for benchmarking machine
  learning algorithms.
\newblock {\em arXiv preprint arXiv:1708.07747}, 2017.

\bibitem{clanuwat2018deep}
Tarin Clanuwat, Mikel Bober-Irizar, Asanobu Kitamoto, Alex Lamb, Kazuaki
  Yamamoto, and David Ha.
\newblock Deep learning for classical japanese literature.
\newblock {\em arXiv preprint arXiv:1812.01718}, 2018.

\bibitem{rps}
Laurence Moroney.
\newblock Rock, paper, scissors dataset, feb 2019.

\bibitem{UCI}
M.~Lichman.
\newblock {UCI} machine learning repository, 2013.

\bibitem{hull1994database}
Jonathan~J. Hull.
\newblock A database for handwritten text recognition research.
\newblock {\em IEEE Transactions on pattern analysis and machine intelligence},
  16(5):550--554, 1994.

\end{thebibliography}

\end{document}